# Enhancement of Healthcare Data Transmission using the Levenberg-Marquardt Algorithm

Qi An[1] and James Jin Kang[1, 2]

[1] Computing and Security, School of Science, Edith Cowan University, Joondalup, WA, 6027 Australia

[2] Corresponding author

e-mail: angela.an{james.kang}@ecu.edu.au

This paragraph of the first footnote will contain support information, including sponsor and financial support acknowledgment. For example, "This work was supported in part by the U.S. Department of Commerce under Grant BS123456."

**ABSTRACT** In the healthcare system, patients are required to use wearable devices for the remote data collection and real-time monitoring of health data and the status of health conditions. This adoption of wearables results in a significant increase in the volume of data that is collected and transmitted. As the devices are run by small battery power, they can be quickly diminished due to the high processing requirements of the device for data collection and transmission. Given the importance attached to medical data, it is imperative that all transmitted data adhere to strict integrity and availability requirements. Reducing the volume of healthcare data and the frequency of transmission will improve the devices' battery life via using inference algorithm. There is an issue of improving transmission metrics with accuracy and efficiency, which trade-off each other such as increasing accuracy reduces the efficiency. This paper demonstrates that machine learning can be used to analyze complex health data metrics such as the accuracy and efficiency of data transmission to overcome the trade-off problem using the Levenberg-Marquardt algorithm (LMA) to enhance both metrics by taking fewer samples to transmit whilst maintaining the accuracy. The algorithm is tested with a standard heart rate dataset to compare the metrics. The result shows that the LMA has best performed with an efficiency of 3.33 times for reduced sample data size and accuracy of 79.17%, which has the similar accuracies in 7 different sampling cases adopted for testing but demonstrates improved efficiency. These proposed methods significantly improved both metrics using machine learning without sacrificing a metric over the other compared to the existing methods with high efficiency.

## I. INTRODUCTION

Recently, the rise of smart cities [1], with their mixture of networks, protocols and devices has forced changes to traditional processes in the healthcare and medical industry. For example, mobile health (mHealth) is a prevalent technology using cloud computing, deep learning, artificial intelligence, big data, and machine learning. Health data are collected from patients' wearable sensor devices and forwarded to the hospital database through technologies supporting cellular networks, and then transmitted to cloud storage systems. The collected medical data is then used for further analysis for medical purposes [2]. Therefore, how to transmit the patients' critical data accurately and efficiently has become a hot topic in the mHealth area. The results of a lack of sufficient and accurate data could bring severe consequences to the patients' diagnosis or treatment and constrain the benefits of mobile technology [3]. Literature reviews showed some work on networking security [4], power consumption [5], and data traffic on sensor devices [6]. Also, some works have been done using machine learning for disease detection, pattern recognition, and medical image processing from gathered medical data [2 - 11]. There are limited studies on both data transmission accuracy and efficiency, which needs further research. Author, James Kang's studies [12 - 16] have focused on improving the trade-off ratio between data accuracy and efficiency using a multilayer inference algorithm. For example, when data accuracy is improved by using a larger quantity of samples, efficiency suffers. Very few studies have provided evidence on machine learning algorithms to improve healthcare data accuracy and efficiency on the network at the same time. Therefore, this paper investigates and evaluates the possibility and feasibility of machine learning method to enhance the healthcare data metrics. The machine learning time series neural network LMA is selected to find out whether this algorithm can enhance healthcare data metrics whilst minimizing the trade-off between accuracy and efficiency.

## II. LITERATURE REVIEW

Machine learning is a sub-field of artificial intelligence that has been used in finance, entertainment, spacecraft, pattern recognition, computer version, computational biology, and medical applications [7]. The increased adoption of machine learning applications in healthcare provides medical experts with advanced capabilities to diagnose and treat disease [8]. Machine learning can support the extraction of relevant data from many patients' datasets stored in electronic health records [9]. Using related disease-causing features from electronic health records, machine learning algorithms can help detect diseases by evaluating data and predicting the underlying causes of the illness [10]. Machine learning provides remarkable advantages for the evaluation and assimilation of amount of complex health data. Furthermore, machine learning can be deployed for data classification, prediction, and clustering compared to the conventional biostatistical approach [11].

Machine learning can be categorized into two types of techniques as supervised learning and unsupervised learning. Supervised learning machine learning trains the algorithms on known input and output data so that it can predict future outputs. Unsupervised machine learning discovers hidden patterns or internal structures in the input data. Supervised machine learning can perform both classifications and regression tasks whist unsupervised machine learning tackles the clustering task [17].

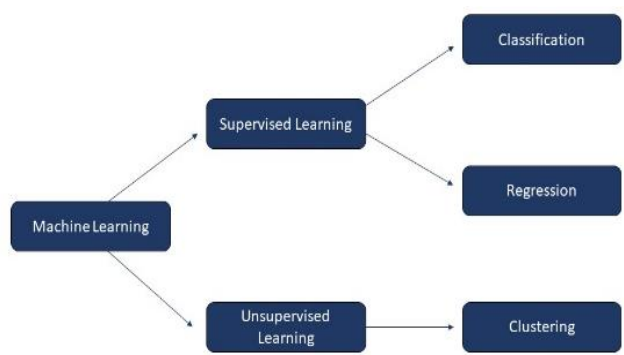

Figure 1: Demonstration of machine learning

### A. SUPERVISED MACHINE LEARNING

#### (1). COMMON SUERVISED CLASSIFICATION MACHINE LEARNING ALGORITHMS

Supervised machine learning classification techniques are algorithms that predict a categorical outcome called classification, the given data labelled and known compared to unsupervised learning. The input data is categorized into training and testing data [18]. The classification algorithms predict discrete response by classifying the input data onto categories. The classical supervised machine learning applications are heart attack prediction, medical image processing and speech recognition [17]. Supervised learning derives classification models from these training data. These models can be used to perform classification on other unlabeled data, the training dataset includes an output carriable that needs to be classified. All algorithms learn specific patterns from the training data and apply them to the test data for a classification problem [17]. Some well-known supervised classification machine learning algorithms are decision tree, support vector machine, naïve bayes, k-nearest neighbors (k-NN), AND neural networks.

1. Decision tree

The decision tree classifier uses graphical tree information to demonstrate possible alternatives, outcomes, and end values. It is a computational process to calculate probabilities in deciding a few courses of action [19]. The decision tree algorithm starts with training data samples and their related category labels. The training set is recursively divided into subsets based on feature values, so the data in each subset is purer than the data in the parent set. Each internal node of the decision tree represents a test feature, every branch node present s the test results, and the leaf nodes present the class label. Since the classifier decision tree is used to identify an unknown sample's category label, it will track the path from the root nodes hold the sample's category label [20]. The advantage of the decision tree algorithm is fast and simple, there is no requirement for any domain knowledge or parameter setting, and high dimensional data can be handled in the context. Also, decision tree algorithms support incremental learning, and it is immutable because of the alternative functions based on each internal node [20]. The disadvantage of the decision tree is that it needs a long training time as it needs to pass over each node level. Also, there is no sufficient memory when it deals with a large database. More errors appear when there are more categories [21].

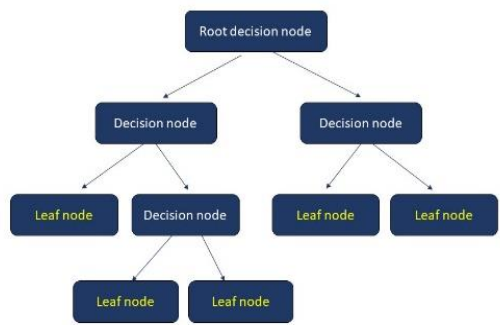

Figure 2: Demonstration of decision tree

Data mining techniques are widely used to help medical professionals diagnose heart diseases. The decision tree is one of the successful machine learning algorithms for heart attack detection in terms of sensitivity, specificity, and accuracy [22]. The decision tree classification algorithm is significantly used to detect and prevent heart disease in the medical industry. Pathak & Valan applied a decision tree to predict heart disease with 88% accuracy through 8 patients' data attributes [23]. Mohamed et al used a decision tree algorithm to reduce data volume by transforming data into a more compact form to save the essential features and provide accuracy in mobile health technology [19].

2. Support vector machine (SVM)

Support Vector Machine (SVM) is a classical technique of machine learning that can address classification problems. Importantly, SVM supports multidomain applications in a big data miming environment [24]. It uses some model features to train the data to generate reliable estimators from the dataset [25]. The concept of SVM is to maximize the minimum distance from the hyperplane to the nearest point of the sample [26].

The advantage of the SVM has a higher performance when dealing with a large dataset than other pattern recognition algorithms, such as Bayesian Network or Neural Network, etc. Also, the main advantage of SVM is that data training is comparatively easy [27]. Most importantly, according to [28], SVM provides a high accuracy among other machine learning algorithms.

The disadvantage of SVM is that it is exceedingly slow in machine learning as a large amount of training time is needed. Memory requirements increase with the square of the number of training examples [28].

SVM is one of the best machine learning algorithms for pattern recognition. Most SVM applications are used for facial recognition, illness detection and prevention, speech recognition, image recognition and face detection [29]. A few researchers applied an optimized stacked SVM to medical application for the early prediction of heart failure (HF). The results revealed that this model has a better performance, which achieved an accuracy range between 57.85% and 91.83% [30]. A study used a fuzzy support vector machine to diagnose coronary heart disease. The experimental results further show that, compared with non-incremental learning technology, this method effectively reduces the computation time of disease diagnosis [31].

3. Naïve bayes (NB)

Naïve Bayes is one of the most widely used classification algorithms. The assumption of Naïve Bayes only includes one parent node and a few independent child nodes [32]. It is the simplest Bayesian network [33]. Naïve Bayes (NB) uses the probability classification method by multiplying the individual probability of each attribute-value pair. This Simple algorithm presumes the independence between attributes and provides remarkable classification results [34].

The strength of the Naïve Bayes algorithm is that it has a short computational data training time [35]. The classification performance can be improved by removing irrelevant attributes [36]. It performs better for the small dataset and can deal with multiple tasks of classification. Besides, it is suitable for incremental training (that is, it can train supplementary samples in real-time) [37]. Not very sensitive to missing data, and the algorithm is relatively simple, often used for text classification; Naïve Bayes is easy to understand the interpretation of the results [38].

The drawbacks of the Naïve Bayes are less accuracy rates than other sophisticated supervised machine learning algorithms, such as ANNs. It requires many training records to achieve excellent performance results [39].

Since Naïve Bayes is very efficient and easy to implement, it is commonly used in text classification, spam filtering or news classification [40]. In the medical field, the Naïve Bayes algorithm is used for disease detection and prediction. A study deployed a Naïve Bayes classifier to skin image data for skin disease detection. It revealed that the results outperform other methods with accuracy from 91.2% to 94.3% [41]. Gupta et al. used Naïve Bayes for heart disease detection through feature selection in the medical sector. The experimental result achieved 88.16% accuracy in the test dataset [42].

4. K-Nearest Neighbors (K-NN)

K-nearest neighbors (K-NN) classification algorithm is one of the simplest methods in data mining classification technology. The assumption of K-NN is to identify the unknown pattern by assigning a value to the K, the nearest neighbour category of the K training samples, is considered as same as the classification [43]. There are few factors involved in the classifier, such as selected K-value and distance measurement and so on [9]. K-NN requires less computational time for training the data than other machine algorithms. However, more computational time is needed in the classification phase [28].

The advantage of K-NN is that it is easy to understand and implement for classification. Besides, it can perform well with many class labels for the dataset. Also, the data training stage is faster than other machine learning algorithms [28].

The drawbacks of the K-NN are computational costs with a sizeable unlabelled sample and time delay during the classification phase. It lacks the principles to sign a K's value, apart from cross-validation, and is expensive computationally. Confusion may occur if there are too many unrelated attributes in the data, leading to poor accuracy. [28].

K-NN is also widely used for disease diagnosis and detection [44]. Researchers attempted to use several data mining techniques to assist medical professionals in diagnosing heart disease. K-NN is one of common data mining techniques in classification problems [45]. For example, some researchers proposed a novel algorithm that combines the K-NN and genetic algorithms to diagnose heart disease, and the proposed method aims to improve predictive accuracy [44]. Shouman et al. investigated whether K-NN can enhance accuracy by integrating other algorithms in diagnosing heart disease. The results found that applying K-NN could achieve a higher accuracy rate than a neural network for heart disease diagnosis [45].

TABLE 1: Comparison of supervised machine learning algorithms are used for classification task

| Classification Algorithms | Decision tree | Support Vector Machine (SVM) | Naïve Bayes (NB) | K-nearest neighbours (K-NN) |
|---|---|---|---|---|
| Prediction accuracy | Low accuracy | High accuracy | Lowest accuracy | High accuracy |
| Learning speed | High speed | Low speed | Highest speed | Highest speed |
| Sample size | Small sample | Large sample | Small sample | Small sample |

### (2). COMMON SUERVISED REGRESSOIN MACHINE LEARNING ALGORITHMS

Supervised regression techniques are algorithms that can predict a continuous response, known as regression techniques [46]. The goal of the supervised regression task is to forecast the outcome's specific value rather than classify the data [47]. The input data are split into training and testing data, the continuous response or target outcome are predicted by selected algorithms [18]. The typical regression techniques are used in algorithmic trading and electricity load forecasting [46]. The popular regression machine learning algorithms are linear regression, logistic regression, ensemble methods, and support vector regression (SVR), and neural networks which are discussed below.

### 1. Linear regression

The linear regression technique is the most simple and desired method to measure the relationship between response variables and continuous predictors. The assumption of the linear regression assumes that the predictors variable and target variables have a linear relationship. The nature of simplicity makes the linear regression technique the best option for small sample analysis with high accuracy, and it is comparatively easy to understand and interpret. However, the model will fail and not achieve the expected result with too many predator variables [48]. Also, it is a one-to-one relationship between variables, and it is not a good fit when dealing with non-linear relationship data [49]. Most problems involve non-linear characteristics to different extents. Linear regression is unsuitable for highly non-linear problems when the relationship cannot be approximated by a linear function between input and output variables. However, before applying other complex machine learning algorithms, it is always worth trying linear regression or other simple machine learning algorithms to understand the difficulty of the problem [50].

### 2. Logistic regression

Unlike linear regression used to predict continuous quantities, logistic regression is mainly used to predict discrete class labels. Logistic regression algorithm predicts the probability with two possible categories for classification problems. Logistic regression uses a logistic function to classify the label in a binary outcome between 0 and 1. Therefore, the output variable can be used to indicate which category the sample belongs to [50].

### 3. Ensemble methods

The ensemble methods are not a single machine learning algorithm, and they gather the strength of other algorithms. They can complete learning by constructing and combining multiple machine learning devices. The advantage ensemble methods have the high predictive accuracy that can be achieved in machine learning, but the model's training process may be more complicated, and efficiency is not possible. There are two standard ensemble learning algorithms currently: Bagging-based algorithms and Boosting-based algorithms. Bagging-based representative algorithms include random forest, and Boosting-based representative algorithms include Adaboost, GBDT, and XGBOOST [51].

There are a few advantages of using ensemble methods. Firstly, it can avoid the overfitting problem. A single machine learning algorithm can easily find many different hypotheses that can ideally forecast all the training data with less accuracy prediction for unseen examples when using the small data size. Thus, using combined algorithms (the different hypotheses of Averaging) minimizes the risk of selecting unsuitable hypotheses, thus improving overall forecasting performance. Secondly, ensemble methods have the advantage of computation. Ensemble methods can reduce the risk of reaching a local minimum by combining several algorithms as a single algorithm perform a local search that may fall into the optimal solution. In any single model of an algorithm, the optimal hypothesis may go outside of space. Ensemble methods can extend the search space so the fit space the data by integrating different algorithm models. The ensemble methods suit complex problems with large data sets [52].

### 4. Support vector regression (SVR)

Support vector regression (SVR) is a supervised regression technique used to study the relationship between one or more independent variables and a dependent variable (continuous value). Unlike linear regression techniques that rely on model assumptions, SVR learns the importance of variables to characterize the relationship between input and output [53]. SVR attempts to seek a linear regression function that can maximally approximate the vector of actual data output with tolerance of error [54]. Different factors will influence the generalization ability of SVR. For example, the unseen data in the after-training data set is predicted. Hence, in SVR, some parameters need to be adjusted with sigmoid and radial basis, polynomial and the kernel function-linear, the kernel function's gamma, the number of required polynomial kernel function, kernel function's bias, error penalty parameter, and radius. The adjustment of those parameters is to make sure the generated model avoids overfit or underfit problems with the data [55].

One of the primary advantages of SVR is that its complexity of computation does not rely on the dimensionality of the input variables. In addition, it has incredible generalization ability and high prediction accuracy [56]. However, SVM is expensive computationally, and it requires a large size of data set.

### 5. Neural networks

Artificial Neural Network (ANN) is such a computational intelligence model. An artificial neural network is an excellent calculation tool based on input parameters and small experimental data with potential output. It optimizes the human brain and self-learning and predicts the output independent of the provided input [57]. A neural network (NN) can perform many regressions or classification functions, although usually only one can be performed per network. Therefore, in most cases, there is a single output variable in the network. However, this may correspond to multiple output units (the post-processing stage is responsible for the mapping output units and output variable [39]. The neural network is calculated by connecting a large number of artificial neurons. The mathematical model or calculation model of the sum function is used to estimate or approximate the function. In most cases, the artificial neural network can change the internal structure based on external information. It is an adaptive system with a learning function in a concept [57].

The neural network's advantage is that it has an absolute charm with regression and statistical models [58]. ANN is the most appropriate prediction algorithm and has a high prediction accuracy compared to other machine learning algorithms [57]. However, one of the main disadvantages of ANN is that it requires a lot of computational work due to its iterative feature [59].

Shi & He [60] states that ANN has intensively used for medical image processing. According to their survey, there are over 33000 items on the medical research topics of image processing over the last 16 years. The majority of research topics are medical image segmentation, medical image pre-processing and medical image object recognition and detection.

TABLE 2: Comparison of supervised machine learning algorithms are used for regression task

| Classification algorithms | K-Means | K-Medoids | Hierarchical | Fuzzy c-Means |
|---|---|---|---|---|
| Prediction accuracy | High accuracy | High accuracy | Low accuracy | High accuracy |
| Learning speed | Fast speed | Fast speed | Low speed | Low speed |

## B. UNSUPERVISED MACHINE LEARNING

Unsupervised machine learning technique is used to analyze the large amounts of unlabeled data with highly non-linear learning using millions of parameters of complex models [61]. It is a common clustering learning technique for exploratory data analysis to group or find hidden patterns in data. It draws inferences from datasets, including input data without labelled responses. Most unsupervised learning applications are used for market research, gene sequence analysis, and object recognition [62]. One of the fundamental rules of unsupervised learning is grouping data into suitable groups. The formal method and algorithms of clustering and grouping the data are based on the object's similarities and characteristics, and the same characteristics are a clustering analysis. It does not classify labels with category, data clustering is distinguished by lack of category information [61]. Clustering algorithms are falling into two groups – hard clustering and soft clustering. Hard clustering occurs when data points belong to one cluster while data points belong to one or more clusters refer to soft clusters. Some popular unsupervised machine learning algorithms are discussed below.

### (1). COMMON HARD CLUSERING UNSUPERVISED MACHINE LEARNING ALGORITHMS

### 1. K-Means

K-Means is an extensively used unsupervised algorithm [61] due to its simplicity and fast speed [63] to solve well-known clustering problems [64]. K-Mean algorithm partitions data points into k clusters by minimizing the sum of squared distance between the point [65] and its nearest neighbour set distance [66]. The matching degree between a point and the cluster depends on the distance from the point to the cluster centre [67]. The best use of the K-Mean algorithm is when the number of clusters is known for fast clustering with a large number of data sets [64]. Therefore, K-Mean remains the most famous population for massive datasets analysis in unsupervised learning [67].

In practice, the K-Means algorithm's advantages have a nature that is easy to learn, fast training speed and has no requirement of following the order of input data, and its "Vector quantization" concept can be used to construct a feature [66].

It can adjust the cluster membership for unsupervised clustering learning tasks [68]. The K-Means have the disadvantages of sensitivity to outlier and scale of data set, the requirement of specifying the number of clusters in advance, resulting in different outcomes with different initial centroids, and inability to handle density and varying size of the convex clusters [68].

2. K-Medoids

K-Medoids is similar to K-Means, but it uses the actual object to find the most central object within the cluster and assign the nearest object to the medoids to create a cluster instead of using the mean value of an object in the cluster as a reference point. K-Medoids is less sensitive to outliners and can adjust the cluster membership. It has a similar limitation of producing different results with different initial centroids. It is best practice when scaling to large data sets, fast clustering of categorical data, and the number of clustering is known [68].

3. Hierarchical clustering

Hierarchical cluster analysis (HCA), also called hierarchical clustering, is a typical cluster analysis method in data mining, which attempts to establish a hierarchical structure of clusters by analysing similarities of the characteristics into clusters [69]. Hierarchical clustering technique recursively produces nested sets of clusters in a dendrogram with the clusters [68]. Two strategies are used in hierarchical cluster analysis, agglomerative and divisive strategy. The Agglomerative clustering strategy approach is known as "bottom to up", it directs "the leaves" to "the root" of a cluster tree. While divisive clustering approach refers to "top down" which directs "the root" to "the leaves". All observations are initially treated as one cluster and then splits are taken place when moving down in the Hierarchical structure [69].

**(2). COMMON SOFT CLUSERING UNSUPERVISED MACHINE LEARNING ALGORITHMS**

1. Fuzzy c-Means

The fuzzy c-Means clustering algorithm is the popular one that clusters the data points when it belongs to more than one cluster. It is also similar to the K-Means, but it is suitable for pattern recognition when clusters overlap. The strengths of fuzzy c-means allow cluster assignment flexibility. It is more practical by being able to provide the probability of belonging to a cluster. However, it has some weaknesses of high complexity in specifying the number of clusters in advance (Govender & Sivakumar, 2020).

2. Gaussian mixture model

The Gaussian Mixture Model (GMM) is an extension of a single Gaussian probability density function [70], which uses multiple Gaussian probability density functions (normal distribution curves) to quantify the distribution of variables accurately. It decomposes the variable distribution into several Gaussian probability of statistical model of densities function (normal distribution curve) distribution [71]. Gaussian mixture model assigns a few single Gaussian distribution, and each of the Gaussian distributions are known as a component with its own evaluation index – covariance and mean. The model adjusts the means, coefficients, and covariance through sufficient number of Gaussian distributions to approximate any continuous function of density closely [72]. Gaussian mixture model can effectively capture the internal correlation structures within data sets [73]. When data points come from different multivariate normal distributions with specific probabilities and belong to more than one cluster, clustering based on Gaussian mixture is partition-based [70]. The Gaussian mixture model is a flexible model for a wide range of distribution probability [74], the feature of clusters can be a few parameters [68]. In addition, it has high accuracy and real-time implementation [75]. The drawbacks of the Gaussian mixture model are computationally expensive with large distributions or few observed data points in data sets. It is difficult to estimate the number of clusters, and it requires large data sets [73].

3. Hidden Markov model

The Hidden Markov model belongs to clustering model-based, and it is useful and suitable for time series data. Each data point represents the observer value according to the time sequence by using the hidden Markov model. Future values are clustering based on the past value (observed value) of the time series. The hidden Markov model includes two sections. The first section is the time series observation that generates the observation, followed by the second section-unobserved state variables [68].

A set of states features the model, and the state-related probability distribution manages to generate time series data. The related state is the first stage- the initial probability distribution, the transition probability matrix connecting successive state, and the dependent probability distribution state. Observers can only see time series observations, while state variables are hidden. The Hidden Markov model provides statistical information such as the standard deviation, mean, and weight value of the cluster based on the cluster's observation results. The Hidden Markov model has the capability of dealing with a variety of types of data. However, this algorithm requires many parameters, and it is suitable for large data sets [68].

TABLE 3: Comparison of unsupervised machine learning algorithms are used for clustering task

| Classification algorithms | Decision tree | Support Vector Machine (SVM) | Naïve Bayes (NB) | K-nearest neighbours (K-NN) |
|---|---|---|---|---|
| Prediction accuracy | Low accuracy | High accuracy | Lowest accuracy | High accuracy |
| Learning speed | High speed | Low speed | Highest speed | Highest speed |

## C. EVALUATION MATRIX FOR MACHINE LEARNING ALGORITHMS

1. Evaluation matrix for supervised classification algorithms

The performance of supervised classification algorithms is commonly evaluated by indicators such as accuracy, sensitivity, specificity. Accuracy is assessing the percentage of predation rate in the model; sensitivity is the amount of the true positive data points are identified correctly in actual positive data points, and specificity is the quantity of the true negative data points are identified in actual negative data points [76].

Accuracy = (true positive data points + true negative data points) / total predicted data points

Sensitivity = true positive data points / actual positive data points

Specificity = true negative data points / actual negative data points

2. Evaluation matrix for supervised classification algorithms

The MSE (Absolute squared error), RMSE (Root Mean Square error), MAE (Mean absolute error), and MAPE (Mean absolute percentage error) are widely used for evaluating the performance of regression algorithms. In addition, the coefficient of determination, known as the R value, is significant for the prediction of accuracy as a valuable indictor in regression tasks [77].

It is crucial to evaluate the performance of clustering algorithms as it is part of data analysis. In supervised learning, the evaluation matrix has developed well and is widely accepted. Unlike supervised learning, its evaluation matrix has not developed well, so it is not easy to define the performance of algorithms. However, some indicators can be used to assess the quality of the model. SSE (Sun of Squared Errors) can be used to calculate the Euclidean distance. The smaller SSE value means a good cluster performance. Calinski-Harabaz index is also called variance ratio criterion, a metric based on dispersion within clusters and between clusters. Silhouette coefficient is used to define the interval with -1 and 1. Rand index and Fowlkes-Mallows Scores (FMI) are used for external criteria validation [78].

This research is to find a suitable machine learning algorithms to enhance healthcare data metrics. Selecting a correct algorithm when using machine learning is significant as there are many machine learning algorithms. After the overview types of machine learning algorithms, supervised machine learning classification algorithms (decision tree, support vector machine, Naïve Bayes, K-nearest neighbours) are not suitable for this study. Supervised classification algorithms attempt to classify or label the data into a discrete outcome. Unsupervised algorithms are clustering algorithms that group or cluster data in a similar group through similar characteristics. Clustering algorithms are also not suitable for the task. This study aims to use machine learning algorithms to forecast the heart rate based on the time series, and it is a regression task to predict a continuous value. Hence, the machine learning regression algorithms technique will be selected to perform the study. However, the linear regression and support vector regression algorithms deal with at least two variables (independent and dependent viable) for a close linear relationship. The heart rata will be predicted according to the time order with a non-linear autoregressive relationship based on the past value. Therefore, they are not the correct approach for the task. Regarding logistic regression, it is like other classification algorithms that can classify the labels. Therefore, it is also not in the range of choice for the regression task. Assemble methods are robust approaches to solving complex problems with large data sets, so the ensemble method is not used for this study. The supervised artificial neural network algorithm can perform regression time series data tasks, especially its outstanding regression and numerical model with high prediction accuracy for large and small data, making it an alternative for this study. A neural network algorithm is selected to conduct the experiment in the next section. Also, after reviewing others' work, there are not many researchers who use machine learning algorithms to predict heart rate on mobile health which focus on the network of wearables or sensor devices. Therefore, it is worth trying the algorithm with different sampling scenarios and seeking the feasibility and acceptability of the proposed method.

## III. METHDOLOGY

### A. DATA COLLECTION

This study uses the healthcare data (heart rate) to conduct the experiment. The University of Queensland Vital Signs Dataset provides open access to researchers and educators. There are 32 case datasets available, which includes the patient's heart rate. Case 9 [79] trend data set is selected with 6550 heart rate data points with 1 hour 49 minutes time duration for this project as the data set is relatively complete compared to the other 31 case data sets.

### B. DATA SPLIT METHOD

Various sampling and training cases were set up to conduct the experiment as shown in TABLE 4 and tested each scenario with LMA to compare and find out the best performer in terms of prediction accuracy and efficiency. The conventional machine learning data division method is 70% to 80% data for training, and the rest is for testing. This study split data into 30% to 90% training data to build the different models and seek the possibility of achieving relatively high accuracy and efficiency to reduce the data size to transmit. It is customary to split the data into training, validation and testing data, respectively. The training data presents the neural network training process, while the validation dataset shows the measurement of generalisation and stopping training. The purpose of validation is to avoid overfitting problems. The testing data is to test the performance of neural network algorithms. It provides an independent measurement of network performance.

TABLE 4: Sampling scenarios of the training dataset

|  | Data split method | | |
| --- | --- | --- | --- |
| Scenarios | Training data | Validation data | Testing data |
| 1 | 90% | 5% | 5% |
| 2 | 80% | 10% | 10% |
| 3 | 70% | 15% | 15% |
| 4 | 60% | 20% | 20% |
| 5 | 50% | 25% | 25% |
| 6 | 40% | 30% | 30% |
| 7 | 30% | 35% | 35% |

### C. ALGORITHM SELECTION

Temporal data, also called time series data, exists in many domains such as medicine, video, and robotics. When managing those types of data, a specific method is needed to consider the samples with a high correlation between consecutive samples in time series [80]. An artificial neural network (ANN) is a computational intelligence model that can perform time series data classification and regression tasks. It optimizes self-learning to predict the output independent of the provided input [57]. The advantage of neural networks is that they have a better performance for regression tasks [58] This study used heart rate datasets over a period of 1 hour 49 minutes with 6312 data points [79]. Heart rates are predicted based on the historical data pattern with a period sequence. Thus, time series neural network algorithms are appropriate for the experiment.

Out of the non-linear autoregressive neural network algorithms, LMA was selected to implement this study. It is the most widely used and most effective nonlinear least squares algorithm for neural network training. Its optimization speed is relatively fast, and it has the advantages of fast convergence speed. It is suitable for simple fitting functions or relatively simple parameters to be estimated. In addition, it mostly guaranteed the algorithm for the accuracy of the function approximation type of problems [81]. MATLAB was used to perform neural networks autoregressive time series - LMA. Hence, the collected heart rates were applied to this algorithm to conduct this experiment.

### D. DATA EVALUATION AND SOFTWAR

The machine learning regression technique is deployed to perform the study. MATLAB R2020b is used for applying machine learning algorithms for the training and prediction of heart rate data. The neural net time series deep learning Toolbox 14.1 is selected to generate LMA.

MSE (mean squared error) and R value from the models evaluate and measure the performance of neural network algorithms. In addition, MAE (mean absolute error) and MAPE (mean absolute percentage error) are calculated after model generation. MAE and MAPE are used to evaluate accuracy (100 – MAPE value). The efficiency can be calculated by (total target data points/trained data points). The 7 scenarios comparison and best-performing scenario for the algorithm is described in the experiment and analysis section.

Formulas were used to evaluate and measure the accuracy and efficiency as below.

$$MSE = \frac{1}{n}\sum_{i=1}^{n}(Y_i - \widehat{Y}_i)^2$$

MAE is the mean absolute error value, which presents the average absolute error between the predicted heart rate and the observed heart rate. It directly calculates the average of the residuals. The lower MAE is, the higher accuracy will be.

$$MAE = \frac{\sum_{i=1}^{n}|Y_i - \widehat{Y}_i|}{n}$$

MAPE is the sum of the average absolute error percentage typically for time series forecasting accuracy or error measurement. The MAPE is calculated by the percentage of mean between the observed heart rate value, and the predicted heart rate value. The lower MAPE means the higher accuracy.

$$MAPE = \frac{1}{n}\sum_{t=1}^{n}\left|\frac{Y_i - \widehat{Y}_i}{Y_i}\right| \times 100\%$$

R value is the coefficient of determination that presents the fitness of the predictive model. The large R squared value indicates that there is less error, and the model is more fit for the prediction. Therefore, the highest R value of the model was selected for accuracy. RSS means the sum of squares of residuals, TSS is the total sum of the squares.

$$R^2 = 1 - \frac{RSS}{TSS}$$

Accuracy of prediction heart rate can be calculated by 100% accuracy minus the mean absolute percentage error value in the model.

$$Accuracy = 100\% - MAPE$$

To save the battery of the sensor device, efficiency is considered by reducing the volume of data. The high efficiency is the less consumption of sensor device battery. It can be calculated by the total amount of data points (heart rates) divided by the training data points (heart rate). $n$ is the total number of heart rates (6312), $t$ presents the total number of training heart rates for each scenario.

$$Efficiency = \frac{n}{t}$$

MSE is known as mean squared error, n presents the total number of data points (6312 heart rates), Yi shows the observed values of heart rate where i is a certain data point, ie.1.2.3.4…n, and (Yi )ˆis the predicted heart rate value. The statistical parameter calculates the average squared of errors of the predicted heart rates and observed original heart rate values. The less error between them, the higher the accuracy.

## IV. EXPERIMENT AND DISCUSSION

LMA is an iterative algorithm used to solve nonlinear least squares problems algorithm [58]. This algorithm combines two least squares algorithms-gradient descent and the Gauss-Newton method. The gradient descent method reduces the sum of the squared errors by updating the paraments in the steepest descent. The Gauss-Newton method reduces squared errors by assuming the least square function is locally quadratic and the parameter estimation value is close to the optimal value range. The combination of the two methods can quickly find the optimal value [81].

TABLE 5 provides information about 7 different training scenarios results from the LMA. The prediction accuracy is very similar in the range of 79% apart from scenario 5. Regarding the least MSE and MAPE, higher R value and accuracy and efficiency, scenario 7 outperforms the other 6 scenarios. The detailed analysis of the scenario is discussed in TABLE 6.

TABLE 5: LMA testing data results

| Training scenarios | MSE | R | MAE | MAPE | Accuracy | efficiency |
|---|---|---|---|---|---|---|
| 1 | 0.20 | 0.9981 | 1.43 | 20.51% | 79.48% | 1.11 |
| 2 | 0.23 | 0.9976 | 1.42 | 20.45% | 79.54% | 1.24 |
| 3 | 0.23 | 0.9975 | 1.42 | 20.47% | 79.53% | 1.42 |
| 4 | 0.24 | 0.9977 | 1.46 | 20.92% | 79.08% | 1.66 |
| 5 | 0.25 | 0.9972 | 1.47 | 21.09% | 78.91% | 2 |
| 6 | 0.23 | 0.9975 | 1.46 | 20.96% | 79.04% | 2.5 |
| 7 | 0.21 | 0.9977 | 1.45 | 20.83% | 79.17% | 3.33 |

TABLE 6 shows the best network performance of the LMA from scenario 7. It uses 30% of samples (1894 heart rates) to train the algorithm, 35% (2209 heart rates) for networking performance validation, and 35% for testing (2209 heart rates).

The lower MSE (0.21) for the test set indicates that the LMA was the best performer of the algorithms tested. Training, validation and testing for R-value are relatively close to 1, especially the testing set result (0.9977) shows the highest value in the entire 7 test scenarios. The MAPE (20.83%) represents the percentage average of the total error data predicted in the test. The model has relatively high accuracy and efficiency with 79.17% and 3.33, respectively.

TBALE 6: Best performance of LMA from scenario 7

| Data split method | Training (%) | Training data points | Validation (%) | Validation data points | Testing (%) | Testing data points |
|---|---|---|---|---|---|---|
| Results | 30% | 1894 | 35% | 2209 | 35% | 2209 |
| MSE | 0.24 | | 0.27 | | 0.21 | |
| R | 0.9975 | | 0.9971 | | 0.9977 | |
| MAE | 1.45 | | | | | |
| MAPE | 20.83% | | | | | |
| Accuracy | 79.17% | | | | | |
| Efficiency | 3.33 | | | | | |

Figure 3 depicts the training and validation MSE (0.27) at 17 epochs. Figure 4 shows the gradient and validation checks result. The gradient decrease in the cost function with the lowest error point from epoch 17 until 23 shows no error increased since then. In addition, the maximum validation checks are increased from epoch 17 to 23, which reaches the default checks at 6, where the errors do not increase with the increasing of epochs.

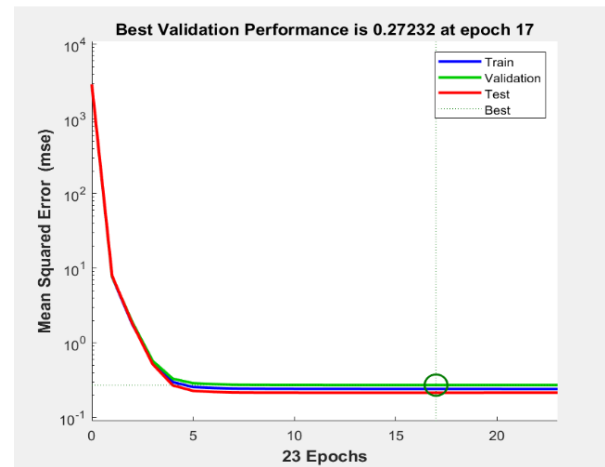

Figure 3: Performance of neural network at epochs 17

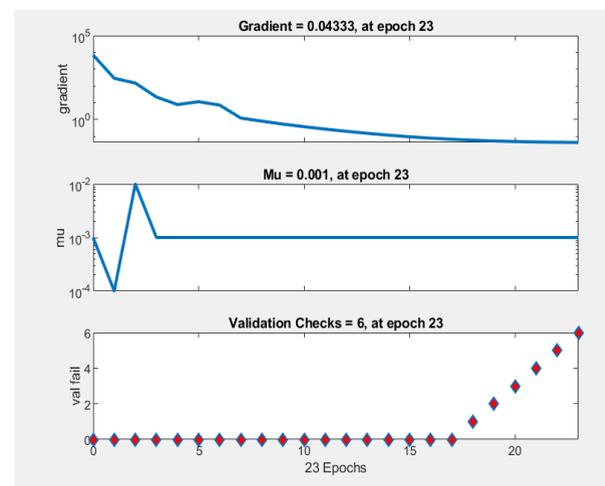

Figure 4: Gradient and validation checks at 6

Figure 5 illustrates the histogram of target – output (heart rate prediction) errors with 20 bins. There are 1700 training data close to zero error, which is 0.024, with 1400 validation data points, and testing is 1800. The fitting tool of the training regression algorithm results can be shown in Figure 6. The training, validation and testing model R value mentioned in Table 2 are all relatively close to 1, which means almost errorless. The model of R value is 0.9974.

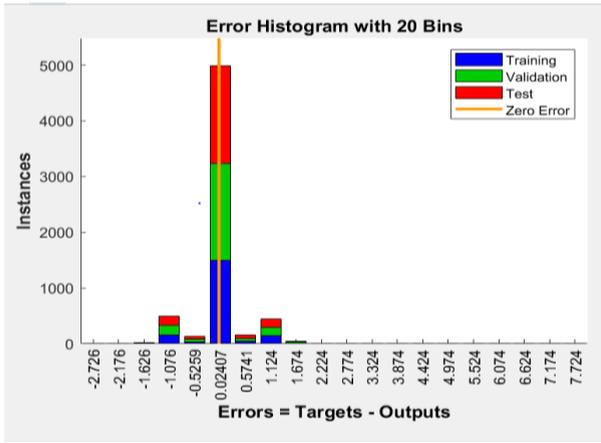

Figure 5: Error histogram with the lowest error (close to 0 error)

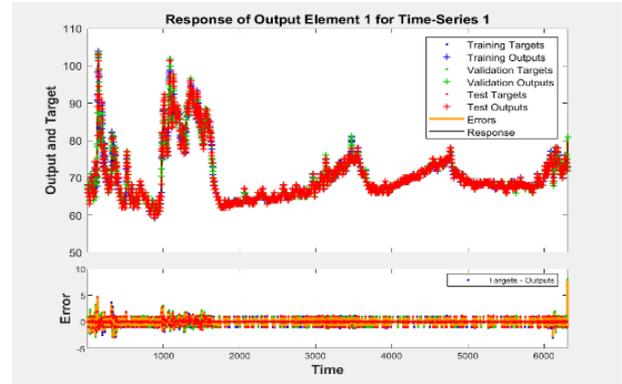

Figure 7: Output heart rate of time series response

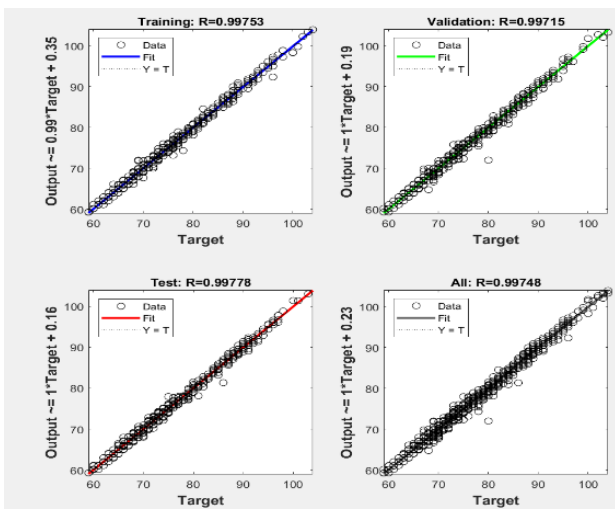

Figure 6: Training regression with R (test 0.997)

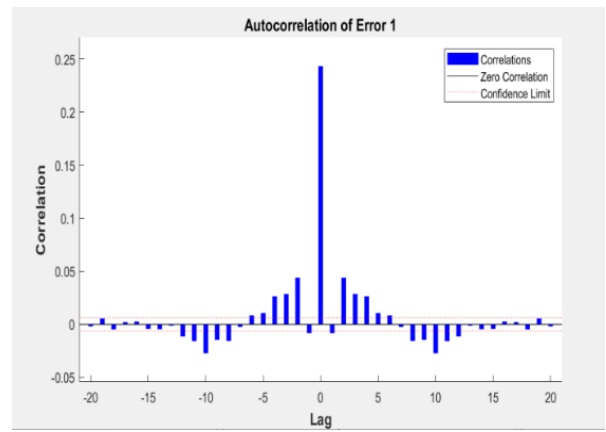

Figure 8: Autocorrelation of error

Figure 7 shows that the time series reflects seconds of an output (heart rate) and errors. Figure 8 demonstrates that the autocorrelation lag 0 is equal to the mean squared error, almost 0.25. It displays the same negative and positive values information symmetrically, indicating that, there is a high autocorrelation, and the prediction errors are related the consecutive heart rate. In addition, the correlation of zero lag means the correlation is within 95% of confidence limits, as such, the model is supposed to be adequate for predicting the accuracy. The experiment used three nonlinear autoregressive LMA across seven different train/validate/test data scenarios and evaluated the best algorithm model in terms of R-value, Mean Squared Error, Mean Absolute Percentage Error, and Accuracy. The experiments in this study showed that the Levenberg-Marquardt neural network algorithm had stable performance in all seven tests. Especially in scenario 7, 30% training data, 35% validation data and 35% testing data are deployed to estimate the neural network performance. Table 8 highlights the least error with MSE (0.21), MAE (1.45), and the R (0.9977).

In 7 tests, the prediction accuracy does not fluctuate in an extensive range with the different data ratios. The results showed that the accuracy is between 70%-80% from the three algorithms. However, efficiency can be a significant improvement without compromising accuracy. Especially Levenberg-Marquardt, the models are quite stable and have a better convergence with all 7 tests during training. It is not an expensive computation and only runs the model with 17 epochs that shows the model is efficient and easy to train with high performance.

For the purpose of comparison, another experiment has been conducted using a different dataset [83]. MATLAB R2021b has been used to apply the LMA, which trained a neural network to predict series y(t) from past values of y(t). The outcome shows that the metrics performance can vary depending on the datasets used for experiments.

Table 7 shows the performance of Levenberg-Marquardt LMA. Heart rate datasets have been used for experiments with training data with 70% and 30% for metrics including R (regression R value), which is the co-relation between outputs and responses, etc. The Mean Square Error (MSE) is deployed to measure the average square of the error between target and output. MSE (Mean Squared Error), R-value, evaluates and measures the algorithm's performance, e.g., R(70) is the outcome of 70% data trained to predict 30% of samples. In addition, MAE (Mean Absolute Error) and MAPE (Mean

Absolute Percentage Error). Table 1 depicts the result of training heart rate data noted with Tr (training), V (validation), Ts (testing), Ds (dataset), Dp (datapoints). Surprisingly, the accuracy is almost the same between targets of training (70) and (30). It indicates that machine learning algorithms perform very well with a lot less of training samples. It affects other metrics such as efficiency, which can be calculated as below.

$$\frac{\text{Number of Total Target Samples}}{\text{Number of Trained datapoints}}$$

TABLE. 7: PERFORMANCE OF LEVENBERG-MARQUARDT SCENARIO.

| Metrics | TrDs | TrDp | VDs | VDp | TsDs | TsDp |
|---|---|---|---|---|---|---|
| Samples 17007 | 70% | 11905 | 15% | 2551 | 15% | 2551 |
|  | 30% | 5103 | 35% | 5952 | 35% | 5952 |
| MSE(70) | 1.868 |  | 1.871 |  | 1.788 |  |
| MSE(30) | 1.534 |  | 1.955 |  | 2.4 |  |
| R(70) | 0.9979 |  | 0.9978 |  | 0.9980 |  |
| R(30) | 0.9984 |  | 0.9976 |  | 0.9974 |  |
| MAE |  | 1.119 (70) |  |  | 1.223 (30) |  |
| MAPE |  | 4.3% (70) |  |  | 4.4 (30) |  |
| **Accuracy** |  | **95.7% (70)** |  |  | **95.6% (30)** |  |
| **Efficiency** |  | **1.43 (70)** |  |  | **3.33 (30)** |  |

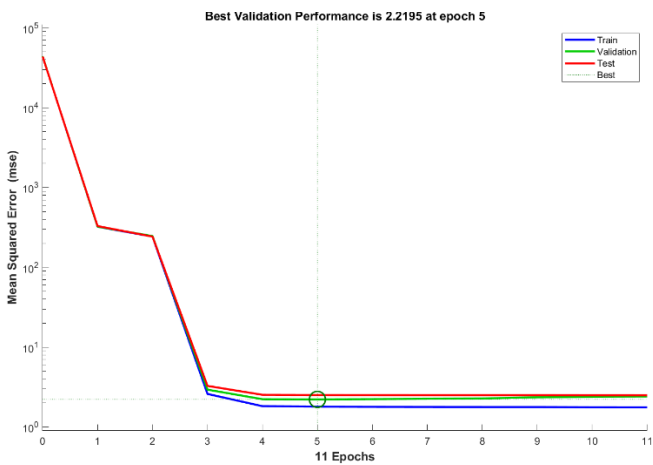

Figure 9: Performance of best validation at Epoch 5, which shows reduced and maintained errors

Figure. 9 shows the best performance from training and validation with an MSE(70) value at 5 Epochs. There is no overfitting that appeared as the error of validation. It halts to increase before the Epoch of 5. The error decreases from Epoch 5 and has a stable movement until the Epoch 11. Fig. 9 compares the best validation performances of MSE (70) and MSE (30) to contrast the MSE outcome, which shows higher errors with fewer training data. The error is reduced significantly with 70% training data comparing to 30% training data as it takes more training to predict with less errors. Comparing the outcome of both training scenarios with 70% and 30%, which are almost identical except the error marked with yellow boxed arrows shows that the algorithm performance improves significantly without much difference for the regression value R whilst the training size is significantly different between 70% and 30% as shown in Table 7. This means that efficiency can improve greatly with little difference in performance metrics such as accuracy. For instance, the efficiency has been improved by 2.3 times whilst maintaining almost the same accuracy (a 0.1% decrease).

## V. CONCLUSION AND FUTURE WORKS

This research applied time series nonlinear autoregressive neural networks using LMA, to predict heart rates. The Levenberg-Marquardt neural network algorithm has the stable performance in the matter of the least error and similar prediction accuracy but with improved efficiency. Therefore, machine learning has proved to improve healthcare data metrics simultaneously compared to the existing methods, which trade-off accuracy and efficiency. It is worthwhile to attempt different data ratios with algorithm and compare the model with a high level of accuracy and efficiency. The performance of metrics may vary depending on the size of the datasets for the mature of machine learning algorithms, and the bigger size of the data may result in a better outcome. Whilst the accuracy doesn't change much, the efficiency can improve significantly. This means that it is feasible to improve efficiency by reducing the sample size with maintaining similar accuracy using machine learning algorithms. This study continues to extend the methods with other healthcare types and size to determine the best metrics by optimizing of training versus prediction samples.

The scenarios depicted in Figure 10 can be achieved when the end-to-end network has been fully implemented from the data capture at personnel devices to the presentation via a smartphone app for the user through data processing and optimization processes by network intelligence with near production approach. The full design and implementation works are under development with human subjects and real-time data collection. It will be a challenge to collect, process and create a personalised health index as it requires accumulated health data for a long period of time with ongoing monitoring for accurate prediction and training performance. Collecting real-time data and transmitting them to a centralized server in military networks are also an area to design and develop for an end-to-end solution, which should consider security and privacy aspects.

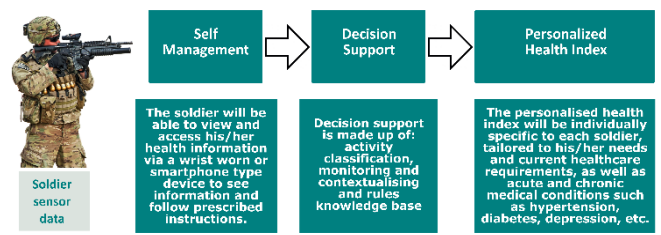

Figure 10: Next stage of Performance Management System (adapted from [82])

Internet of Things (IoT) devices for military purposes can be connected with a server in a military cloud to connect personal sensors devices as well as IoT enabled devices such as weaponry, firearms and communication equipment. In other words, the mHealth based sensor network used for capturing human health data is connected to a military IoT network to collect and transmit data to a centralised server. An example of an IoT-enabled private network is a low powered wide area network (LPWAN), which is typically used in the agricultural industry for crop monitoring. Wireless body area networks (WBAN) are another example of a private network composed of IoT devices which can be integrated with mHealth network. It refers to a collection of inter-networking devices and systems architecture used for optimal collection, classification, and delivery of health information. The journey typically starts from the users/patients (capturing vital physiological signs), traversing across multiple platforms, hops and nodes across the internet as shown Figure 11.

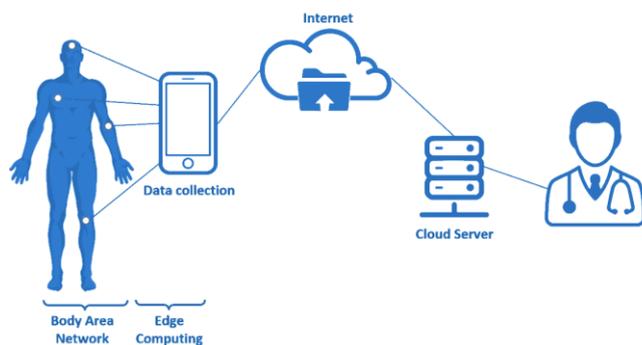

Figure 11: Military network integrated with mobile health, IoT and LPWAN networks (adapted from [14])